\documentclass[letterpaper, 10 pt, conference]{ieeeconf}  

\IEEEoverridecommandlockouts                              

\usepackage{color}
\usepackage{caption}
\usepackage{graphics} 
\usepackage{amsmath} 
\usepackage{amssymb}  
\usepackage{biblatex} 
\usepackage{booktabs}
\usepackage{graphicx}
\usepackage[capitalise]{cleveref}
\title{\LARGE \bf
UNO: Uncertainty-aware Noisy-Or Multimodal Fusion for Unanticipated Input Degradation
}

\author{Junjiao Tian, Wesley Cheung, Nathaniel Glaser, Yen-Cheng Liu and Zsolt Kira\\
Georgia Institute of Technology\\
\texttt{$\{$jtian73,wcheung8,nglaser3,ycliu,zkira$\}$@gatech.edu}
}
\begin{document}

\maketitle
\thispagestyle{empty}
\pagestyle{empty}

\begin{abstract}
 The fusion of multiple sensor modalities, especially through deep learning architectures, has been an active area of study. However, an under-explored aspect of such work is whether the methods can be robust to degradation across their input modalities, especially when they must generalize to degradation not seen during training. In this work, we propose an \textit{uncertainty-aware} fusion scheme to effectively fuse inputs that might suffer from a range of known and unknown degradation. Specifically, we analyze a number of uncertainty measures, each of which captures a different aspect of uncertainty, and we propose a novel way to fuse degraded inputs by scaling modality-specific output softmax probabilities. We additionally propose a novel data-dependent spatial temperature scaling method to complement these existing uncertainty measures. Finally, we integrate the uncertainty-scaled output from each modality using a probabilistic noisy-or fusion method. In a photo-realistic simulation environment (AirSim), we show that our method achieves significantly better results on a semantic segmentation task, as compared to state-of-art fusion architectures, on a range of degradation (e.g. fog, snow, frost, and various other types of noise), some of which are unknown during training. 
\end{abstract}


\section{INTRODUCTION}

Image-based scene understanding methods for robotics, such as object detection and semantic segmentation, have been extensively studied and steadily improved in the past few years. However, robots commonly use multiple sensing modalities beyond simple color imaging, and therefore there is an increasing interest in learning how to leverage this additional sensor information to complement the standard image data. For example, in some cases, depth information can help better separate objects that lack distinguishing textures and colors. 

In general, utilizing multiple modalities entails the fusion of different sensor streams that provide potentially complementary information. For example, depth estimation typically degrades quickly with distance, either in accuracy or resolution. Many works have explored where to fuse modality-specific streams topologically~\cite{schlosser2016fusing, fusenet2016accv, valada2018self}. In general, researchers have attempted different fusion schemes such as early, late and hierarchical fusion schemes.  Many works have also explored fusion schemes at different levels of representation in order to increase the interaction of different modalities~\cite{fusenet2016accv,valada2018self}. 

\begin{figure}
\centering
   \begin{picture}(300,190)
     \put(0,0){\includegraphics[width=\linewidth]{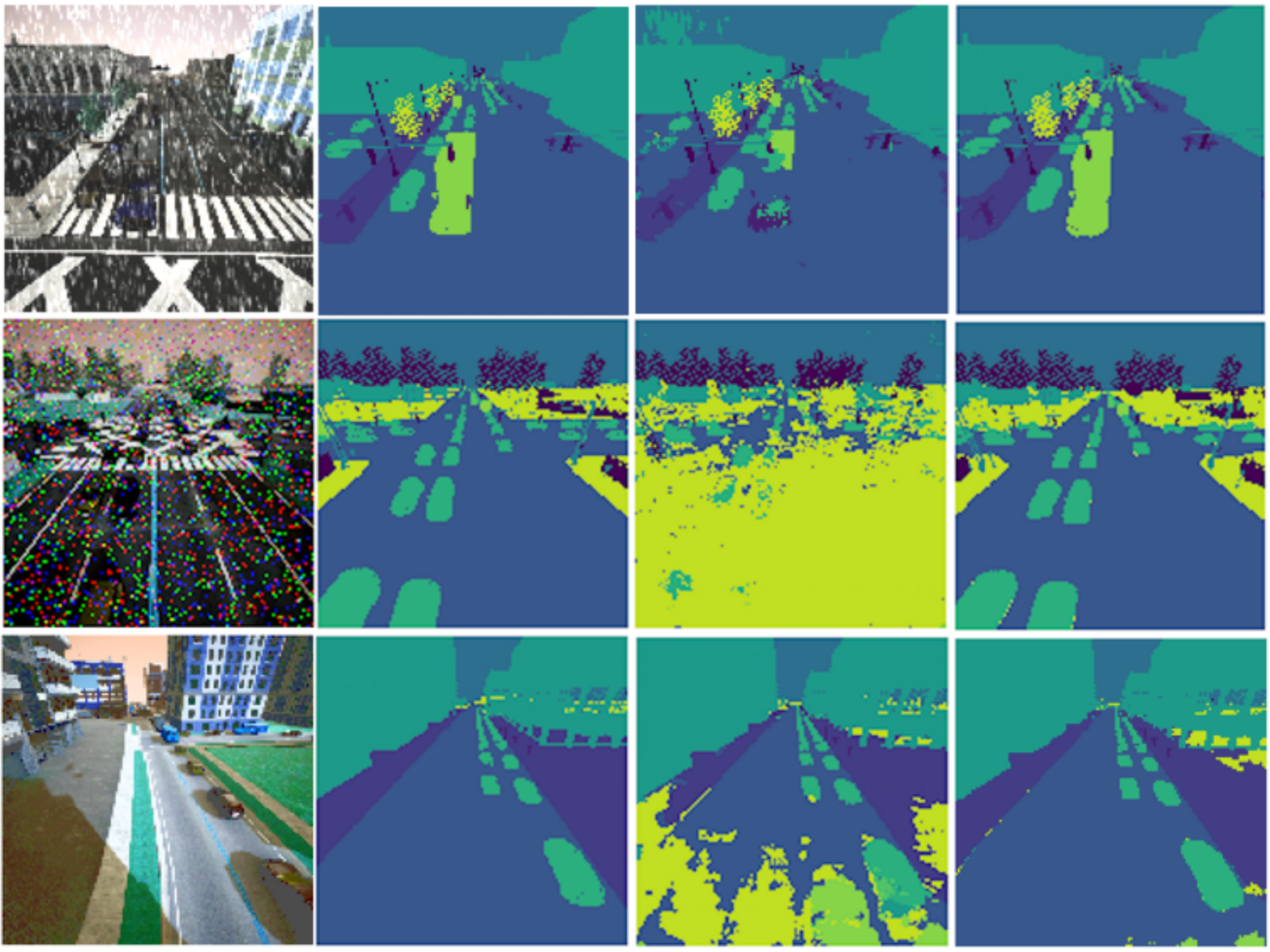}}
      \put(-2,-10){\small{Degraded RGB}}
      \put(85,-10){\small{GT}}
      \put(135,-10){\small{SSMA~\cite{valada2018self}}}
      \put(190,-10){\small{UNO++ (Ours)}}
   \end{picture}
\setlength{\belowcaptionskip}{-15pt}
\caption{\textbf{Performance of state-of-the-art fusion model and ours on degraded RGB data}. First row: Snow. Second row: Impulse noise. Third row: Brightness degradation. Note that the depth channel is not degraded.}%
\label{fig:SSMA_ours}%
\end{figure} 

However, the variety of scenes and degradation in the real world presents a challenge to all fusion schemes. As such, the ability to automatically adapt to a changing environment is the key to safety in application such as robotics and autonomous driving. A robust fusion scheme should dynamically adapt to sensor failure and noise, emphasizing the modalities that are less corrupted and more informative. Various works with different gating and attention mechanisms~\cite{mees2016choosing, valada2017adapnet, valada2018self} have demonstrated the importance of weighting different modalities depending on the scene. Yet, there have been few works on adapting fusion models to a variety of degradation and noise, especially those that do not appear in the training data (i.e. noise that is unknown \textit{a-priori}). 
In this paper, we investigate an adaptive fusion scheme for unseen degradation with application to RGB-D semantic segmentation. We leverage recent development in uncertainty estimation for deep neural networks \cite{gal2016uncertainty,kendall2017uncertainties} and show that different uncertainty measures correlate differently to different types of degradation. We therefore propose a method to combine multiple types of uncertainties by representing their deviation from the training set (\textit{deviation ratio)}, and use this criteria in a novel way to \textit{calibrate} the output prediction probabilities. We further add an additional novel data-dependent spatial temperature scaling that models a spatial type of uncertainty not covered by existing approaches. Given uncertainty-modulated outputs from each modality, we finally propose a simple but flexible uncertainty-aware probabilistic fusion method with no learning parameters and show robust performance across different degradation.

Using a photorealistic simulator (AirSim~\cite{airsim2017fsr}) across a variety of conditions such as fog, snow, and various types of noise, we show that our method achieves stronger fusion results than current state-of-art. Our improvement is especially noticeable in the cases where the specific degradation is not represented in the training set. Fig.~\ref{fig:SSMA_ours} shows examples of performance of the state-of-the-art model SSMA~\cite{valada2018self} and our proposed method.
Our method achieves a relative improvement in mean IoU of 11\% over our strong baselines and 28\% over state-of-art fusion methods.

The contributions of this paper are as follows:
\begin{itemize}
    \item We propose a method for combining several uncertainty metrics, which capture different aspects of uncertainty, using a \textit{deviation ratio} that encodes how the metrics deviate from the training set. 
    \item We introduce an additional uncertainty method,\textit{ the spatial temperature network}, which captures a data-dependent spatial uncertainty that is absent in the existing uncertainty metrics. 
    \item We propose a probabilistic uncertainty-aware fusion scheme,\textit{ Uncertainty-aware Noisy-Or (UNO)}, that dynamically adapts to the changing environment by combining an arbitrary set of experts (e.g. modalities or architectures). Our method has several advantages, including speed (i.e., no training needed) and the ability to dynamically fuse an arbitrary (and potentially changing) set of modalities. 
\end{itemize}

\section{RELATED WORK}
\textbf{Fusion Architectures for RGB-D Semantic Segmentation } A number of fusion architectures have been developed for combining modalities~\cite{schlosser2016fusing, fusenet2016accv, valada2018self}, and recently it has been shown that variations of attention and gating mechanisms~\cite{kim2018gatedfusion, deng2019rfbnet, zeng2019confidencemap} can adapt to dynamic environments by weighting modalities differently for conditions that occur in the training data. While most works~\cite{mees2016choosing, valada2017adapnet} have considered external environmental degradation such as rain, snow, glare, low-lighting, and seasonal appearance changes, more recent works~\cite{kim2019noisychannelrobustness} address robustness to different types of internal image degradation such as Gaussian noise. However, we observe that these methods are both trained and tested on the same subset of degradation--the test set noises are in the training distribution and hence explicitly learn-able by the network. Although \cite{hendrycks2019robustness} includes ways to augment data by applying many common corruptions and perturbations, we posit that it is unrealistic to anticipate every degradation that may be encountered. Hence, our methods attempt to address this key limitation of current work, namely their inability to reliably and feasibly address all possible forms of degradation. For more details on the recent trends and architectures for multimodal fusion, we refer to~\cite{ramachandram2017fusionsurvey}.

\textbf{Noisy-Or Approximate Bayesian Inference } We model the fusion process as Bayesian inference. A major difficulty in obtaining conditional probability from a Bayesian Network is the complexity of completing the conditional probability distribution table (CPT) which grows exponentially with the number of parents. 
Under certain cause-independent assumptions~\cite{heckerman1994new}, the conditional probability in a Bayesian network can be approximated by a Noisy-Or gate~\cite{pearl2014probabilistic}. Unlike logical-or, Noisy-Or is more realistic because each parent has a non-negligible probability of being inhibited. Practically for fusion, this means that no hard threshold is needed and predictions from each modality are considered with non-negligible probability. This framework has been expanded to include a leak probability which accounts for causes not covered by all of the independent parents~\cite{henrion1987some}. \

\textbf{Uncertainty Estimation} We are interested in an uncertainty estimator that correlates well with the degree of anticipated and unanticipated degradation. Regarding unanticipated degradation, there are many works on detecting misclassification and out-of-distribution (OOD) data by measuring certain notions of uncertainty or confidence. It has been noted that out-of-distribution images can be identified through epistemic uncertainties~\cite{kendall2017uncertainties}. Many other per-pixel uncertainty measures have also been developed and compared in~\cite{blum2019fishyscapes} including ODIN~\cite{liang2017odin}, Bayesian networks~\cite{kendall2015bayesian, gal2018bayesiandeeplab}, density estimation~\cite{dinh2016realnvp}, and OOD training~\cite{devries2018OoDconfidence}.

In particular, Bayesian networks~\cite{kendall2015bayesian, gal2018bayesiandeeplab} have been shown to yield desirable properties for modeling uncertainty, such as a model confidence that correlates with accuracy. Their methods involve using Monte-Carlo dropout (MCDO)~\cite{gal2015mcdodropout} as the primary means of approximate inference, with several measures such as predictive entropy and mutual information that can be calculated from these sampling passes.

We show in our scenarios that model uncertainty (approximated by dropout) does not correlate with all degradation, and calibration methods such as temperature scaling~\cite{guo2017calibration}, must be trained on the specific degradation that will be encountered. In our work we propose a novel \textit{uncertainty-based} calibration as well as a data-dependent spatial form of temperature scaling; we then combine multiple notions of uncertainty to maximize robustness.  

\section{METHOD}

\begin{figure*}
\centering
\includegraphics[width=\linewidth]{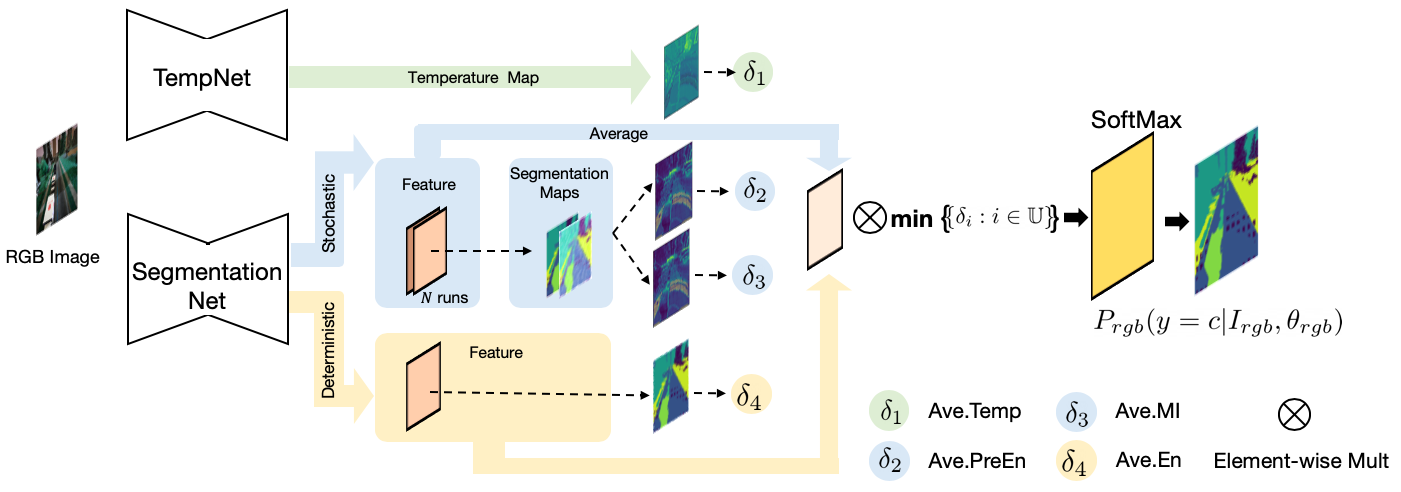}
\setlength{\belowcaptionskip}{-15pt}
\caption{\textbf{Overall pipeline for a single expert (RGB branch)}. We combine multiple uncertainty metrics using the deviation ratios ($\delta$s) calculated for a test sample using different uncertainty metrics. Deviation ratio is defined in~\cref{DeviationRatio}. Either stochastic model or deterministic model is used at a time. The performance of both is examined in~\cref{results}. }%
\label{fig:overall}%
\end{figure*} 

In this section, we introduce the Uncertainty-aware Noisy-Or (UNO) fusion scheme. We briefly highlight three conventional uncertainty metrics, each of which captures a different element of uncertainty, and our proposed \textit{deviation ratio} in~\cref{DeviationRatio} that allows us to combine these uncertainty metrics. We then describe a novel learning-based uncertainty metric, \textit{TempNet} (see~\cref{TempNet}), for capturing a data-conditioned spatial uncertainty that is not covered by the existing approaches. Finally, we propose a probabilistic framework for multi-modal fusion in~\cref{NoisyOr}.

In the discussions below, we define $\mathbb{C}$ as the set of classes we are interested in classifying and $\mathbb{U}$ as the set of employed uncertainty metrics. We view each modality-specific segmentation network as a different expert, and we denote $\{E_1,..,E_i\} \in \mathbb{E}$ as the set of independent segmentation networks (experts).  

\vspace{-0.1cm}
\subsection{Uncertainty Estimation and Deviation Ratio}\label{DeviationRatio}
A number of methods exist for producing uncertainty estimates over the output predictions of a neural network~\cite{guo2017calibration,gal2016uncertainty,kendall2017uncertainties}, namely predictive entropy, mutual information, and deterministic entropy. While many works focus on the accuracy of such estimates, we propose to use them to re-weight modalities during fusion. Unlike hand-chosen weighting methods~\cite{papandreou2007uncertaintyfusion} that use uncertainty, we propose to modulate modalities in a novel way through automatic \textit{scaling} of the outputs scores (and hence softmax probabilities); this procedure is similar to calibration methods~\cite{guo2017calibration} but conditioned on uncertainty. 
The goal of uncertainty scaling is to ``soften'' the softmax probabilities depending on how different (i.e. out-of-distribution) the uncertainties are from the training data. For example, if a modality is corrupted by noise, we want to flatten its distribution to represent maximum uncertainty, allowing our fusion scheme (see Section~\ref{NoisyOr}) to effectively ignore the erroneous outputs from this modality. 

We first describe the existing uncertainty metrics used in this paper, as proposed by~\cite{gal2016uncertainty}. Given a test sample $x$ and training data $D_{train}$, the 
\textbf{predictive entropy} of a predictive distribution $\mathbb{H}[y|x,D_{train}]$ can be approximated by collecting outputs from $T$ stochastic forward passes with different dropout samples through the network (i.e. MCDO):
\begin{align}
    &\hat{\mathbb{H}}[y|x,D_{train}] \approx \nonumber\\
    &-\sum_c^{\mathbb{C}}{ \left( \frac{1}{T} \sum_t{p(y=c|x,\hat{\theta}_t) \log{\frac{1}{T} p(y=c|x,\hat{\theta}_t)}} \right)},
\end{align}
where $c$ is over all classes, $p(y=c|x,\hat{\theta}_t)$ is the probability of class $c$ given input $x$, and $\hat{\theta}_t$ is the sampled weights at stochastic pass $t$.\\
The \textbf{mutual information} can be approximated with a similar procedure:
\begin{align}
    \hat{\mathbb{I}}[y,w|&x,D_{train}] \approx \hat{\mathbb{H}}[y|x,D_{train}] \nonumber\\
    & + \frac{1}{T}\sum_{c,t}{p(y=c|x,\hat{\theta}_t)\log p(y=c|x,\hat{\theta}_t)},
\end{align}
We also compare with the \textbf{entropy} of a deterministic model:
\begin{align}
     \mathbb{H}[y|x,&D_{train}] =\nonumber\\
     &-\sum_c{p(y=c|x,\theta) \log{p(y=c|x,\theta)}},
\end{align}
where $w$ is the model's learned parameters, which are fixed at inference (note only one forward pass is required).

To automatically scale the output probabilities conditioned on the degradation level, a new metric that scales dynamically with uncertainty is needed. 
We propose to capture how deviated a test sample is from the training distribution using a combination of the preceeding uncertainty metrics. We define a deviation ratio, which reports a numeric score less than unity if a test sample is out-of-distribution and unity if the sample is in-distribution (note that, intuitively, degradation should increase uncertainty):

\begin{align}
    \delta = 
    \frac{\mu_{train}}{max \left(0,\mu_{test} - \mu_{train} - \sigma_{train}\right) + \mu_{train}},
    \label{eq:deviation_ratio}
\end{align}
where $\mu_{train}$ is the training average of a specific uncertainty metric aggregated across all images and averaged over pixels in the training set and $\sigma_{train}$ is the standard deviation of the uncertainty metric scores. $\mu_{test}$ is the average uncertainty score for a test sample. The uncertainty metrics provide pixel-wise scores, which we average over an entire image. We perform this averaging step because the per-pixel uncertainty metrics can be unreliable as a local indicator of deviation, as shown by~\cite{blum2019fishyscapes}. Thus, from the three uncertainty metrics listed above, three deviation ratios can be calculated: average mutual information (\textbf{Ave.MI}), average predictive entropy (\textbf{Ave.PreEn}), and average entropy (\textbf{Ave.En}).

 In the end, we combine the deviation ratios of different metrics using a Min operation. Intuitively, we choose the metric that is most sensitive to the current degradation, adopting a conservative selection method in which we assume worst-case for a model's uncertainty: 
\begin{align}
    \delta_{min} = \text{min} \left([\delta_i:i\in\mathbb{U}]\right)
\end{align}
To calibrate a network to reflect uncertainty in the presence of degradation, we define the uncertainty-calibrated prediction:
\begin{align}
    p_i = \text{Softmax}\left( [l_i^1,...,l_i^c] * \delta_{min} \right), 
\end{align}
where $[l^i_1,...,l^i_c]$ are the pre-softmax logits. 
Thus, if a sample is far from the training distribution, $\delta$ will be a scalar less than 1 and thus ``softens'' the distribution and makes the model less confident in its prediction. The complete segmentation and scaling pipeline for the RGB branch is shown in Fig. \ref{fig:overall}. 

For each modality-specific expert, a different $\delta_{min}$ is calculated. For the multimodal expert, we do not extract new deviation ratios; rather, a second Min operation is performed on all the $\delta_{min}$'s from involved modality-specific experts.

\begin{figure}
\centering
\includegraphics[width=7.5cm]{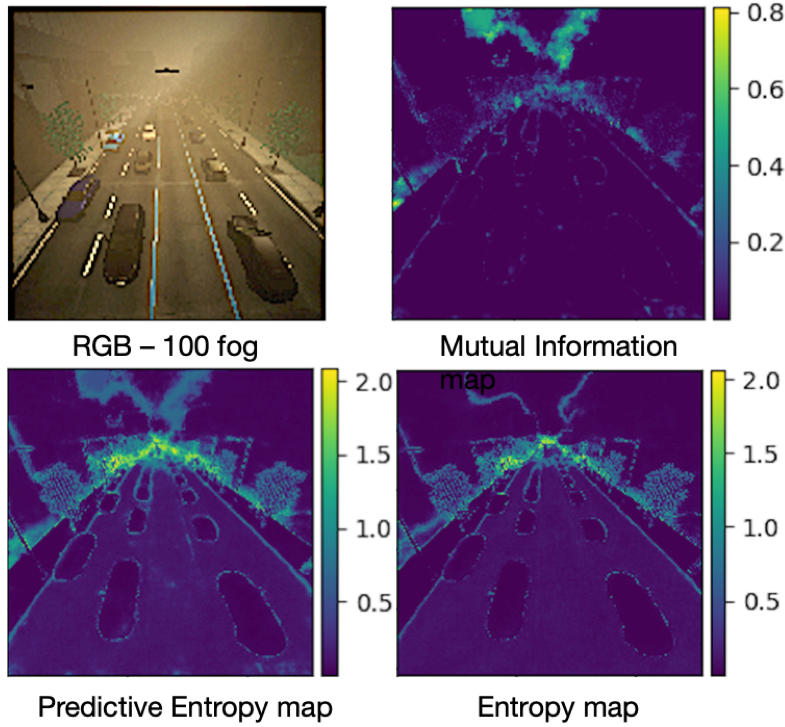}
\setlength{\belowcaptionskip}{-20pt}
\hspace{-0.5cm}
\caption{\textbf{RGB uncertainty measurements with significant fog degradation}. Conventional uncertainty metrics do not capture the spatial  degradation caused by fog.}%
\label{fig:uncertainty}%
\end{figure} 

\subsection{Spatial Temperature Network (TempNet)}\label{TempNet}
In the calibration literature, conventional temperature scaling~\cite{guo2017calibration} is not able to adapt to different test conditions because it utilizes a scalar trained to a specific calibration set. 
Another observation is that degradation is often regional. For example, fog affects vision further into the distance, and a good temperature model should be able to flatten the distributions for points more distant from the viewpoint. 

We therefore introduce a spatial and data-conditioned temperature network to capture uncertainty induced by degradation. We show that the output of this model can be interpreted as an uncertainty which is not captured by conventional uncertainty extraction methods. As an example, Fig. \ref{fig:uncertainty} shows an example of fog degradation and corresponding uncertainty maps. These uncertainties capture uncertainty along edges of objects but do not capture spatial uncertainty.  

TempNet is a shallow version of SegNet and uses 2 convolution and pooling/upsampling blocks for the decoder and encoder. Unlike prior methods, the scores produced by TempNet are conditioned on the data.  The input to the temperature network is the same as for the segmentation network, and the output is a single-channel \textit{spatial temperature map}, $T  \in R^{d_o \times d_o}$. 
The average temperature deviation ratio (\textbf{Ave.Temp}) uses the average of the test spatial temperature map and applying~\cref{eq:deviation_ratio}. 

To train TempNet, we minimize the Negative Log Likelihood of the correct class label for each pixel.
\begin{align}
    p_{ij} = \text{Softmax}\left( [l_1,...,l_c]_{ij} * t_{ij} \right), \\
    L = - \sum_{ij} \log \left( p_{ij}(y=c|x,\theta)\right), 
\end{align}
where $t_{ij}$ is the $i_{th} \text{ row and }j_{th}$ column element of the temperature map $T$ and where $l_{ij}$ is the pre-softmax logit of the segmentation output. 
The segmentation network is pretrained and kept fixed when training the temperature network. The same training procedure is done independently per modality. 

\subsection{Noisy-Or Fusion}\label{NoisyOr}
We now introduce our proposed probabilistic fusion scheme that combines the uncertainty-modulated outputs of each modality. The final predictive distribution for a pixel is obtained by a \textit{Noisy-Or operation} for each class and then normalized across all classes:
\begin{align}
    I(y=c) &= 1 - \prod_i 1-p_i(y=c|x_i,\theta_i) \quad \forall i \in \mathbb{E}\label{eq:NoisyOr_eq},\\ 
    p(y=c) &= \frac{I_c}{\sum_j I_j}  \quad \forall j \in \mathbb{C},
\end{align}
where $p_i(y=c|x_i,\theta_i)$ is the predictive probability of expert $i$ for class $c$, $x_i$ and $\theta_i$ are the input and parameters of expert $i$ and $p_c$ is the final probability for class $c$.

In Bayesian networks, Noisy-Or can be used to model causality between $N$ causes ${E_1,E_2,...,E_N}$ and their common effect $Y$ under certain independent causality assumptions: 1) Each of the causes is sufficient to produce the effect in the absence of all other causes, and 2) The ability of being sufficient is not affected by the presence of other causes.

\begin{figure}
\centering
\includegraphics[width=\linewidth]{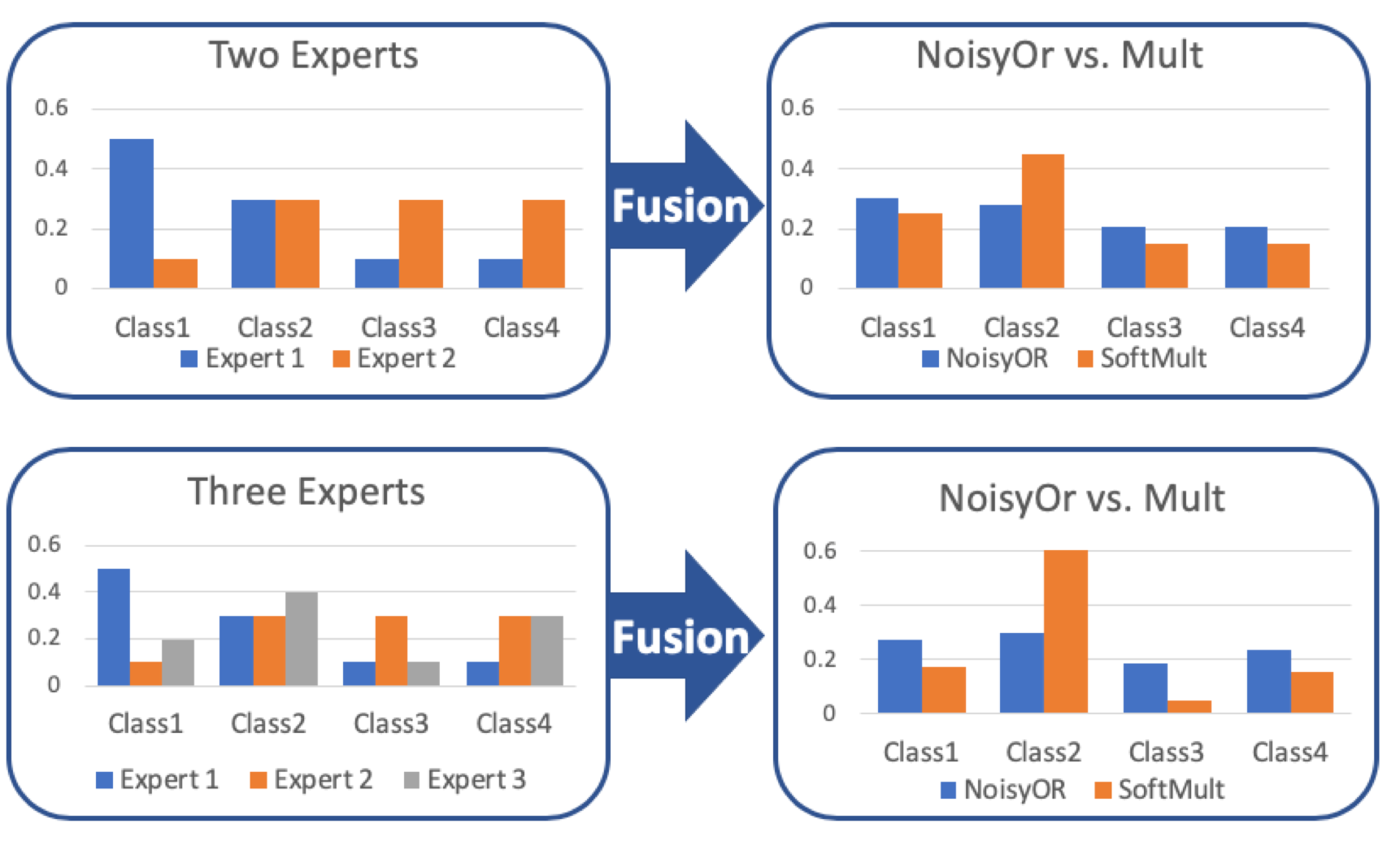}
\setlength{\belowcaptionskip}{-15pt}

\caption{\textbf{Noisy-Or vs. Multiplication 4 class prediction Example}. \textbf{Row 1}: Expert 1 is confident in class 1 while expert 2 is uncertain. Noisy-Or selects case 1 whereas Mult selects case 2 on which the two experts agree the most. \textbf{Row 2}: A third expert gives a confident prediction for class 2, and both Noisy-Or and Mult select case 2.} 
\label{fig:noisyor_vs_mult}%
\end{figure}

Practically, Noisy-Or has some desirable  properties for fusion. When multiple experts give different predictive distributions to some discrete classes, Noisy-Or preserves disagreement and accentuates agreement between experts. Unlike multiplicative fusion which simply multiplies (i.e. smooths) the contributing probabilities, Noisy-Or selects the class on which one expert is significantly more confident. Colloquially speaking, if one expert produces a very confident prediction, the overall prediction should not necessarily be softened by other, less confident experts. Noisy-Or maintains expert-level confidences while still being robust to outliers.A toy example demonstrates the flexibility of Noisy-Or in~\cref{fig:noisyor_vs_mult}. 

We argue that these independent causality assumptions are satisfied because each $E_i$ is a complete segmentation network capable of producing a probability $p_i(y=c|x_i,\theta_i)$ independent of other experts and is not affected by the presence of others. The leak node models the causes not covered by the independent experts. In our case, we adopt an off-the-shelf RGB-D fusion model as the leak node. The causality independence assumption makes this framework flexible. It is easy to add or remove new modality-specific experts, and furthermore, any multimodal expert can be introduced as a leak node. By incorporating multiple uncertainty-aware modality-specific experts and multimodal experts into one framework, our Noisy-Or fusion model is robust to OOD.

\begin{table*}
\resizebox{\linewidth}{!}{%
\begin{tabular}{lrr|rrrrrrr|rrr|r}  
\toprule
&\multicolumn{2}{c}{In Distribution} & \multicolumn{7}{c}{RGB Degradation} & \multicolumn{3}{c}{Depth Degradation}\\
 & 0 fog & 50 fog & 100 fog & \scriptsize{MotionBlur} & \scriptsize{Frost} & \scriptsize{Snow}& \scriptsize{Brightness} & \scriptsize{Blackout} & \scriptsize{Impulse} & \scriptsize{Gaussian} & \scriptsize{ShotNoise} & \scriptsize{Impulse} & \scriptsize{Average} \\
\midrule
1, Ave.Temp & 80.75 & 77.79 & 75.73 & 73.36 & 75.59 & 74.89 & 72.47 & 78.71 & 78.74 & 68.22 & 72.34 & 64.86 & 74.45\\
2, Ave.En & 82.62 & 79.70 & 77.34 & 81.21 & 80.37 & 79.50 & 78.32 & 79.65 & 79.43 & 77.55 & 79.32 & 47.62 & 76.89 \\ \midrule
3, Ave.PreEn & 80.80 & 77.79 & 77.17 & 79.68 & 78.78 & 78.06 &  77.51 & 78.65 & 78.52 & 75.47 & 77.62 & 45.43 & 75.46  \\
4, Ave.MI & 80.81 & 77.80 & 78.07 & 78.28 & 78.14 & 78.05 & 78.30 & 73.41 & 76.69 & 75.76 & 77.62 & 47.84 & 75.06 \\
\midrule
\textbf{min}(1,2)  & 82.62 & 79.69 & 77.67 & 81.24 & 80.38 & 79.57 & 78.33 & 79.72 & 79.43 & 77.55 & 79.32 & 64.80 & \textbf{78.36}\\
\textbf{min}(1,3,4) & 80.77 & 77.75 & 78.14 & 78.29 & 78.35 & 77.94 & 78.09 & 78.53 & 78.52 & 75.79 & 77.64 & 63.53 & 76.95 \\
\bottomrule
\end{tabular}}
\caption{\textbf{Ablation:} performance of Noisy-Or fusion with different deviation ratios (uncertainty metrics) on Mean IoU. Ave.Temp and Ave.En require a single deterministic pass whereas Ave.PreEn and Ave.MI require multiple MCDO passes.}
\label{tab:ablation}

\end{table*}
\begin{table*}
\resizebox{\linewidth}{!}{%
\begin{tabular}{lrr|rrrrrrr|rrr|r}  
\toprule
&\multicolumn{2}{c}{In Distribution} & \multicolumn{7}{c}{RGB Degradation} & \multicolumn{3}{c}{Depth Degradation}\\
 & 0 fog & 50 fog & 100 fog & \scriptsize{MotionBlur} & \scriptsize{Frost} & \scriptsize{Snow}& \scriptsize{Brightness} & \scriptsize{Blackout} & \scriptsize{Impulse} & \scriptsize{Gaussian} & \scriptsize{ShotNoise} & \scriptsize{Impulse} & \scriptsize{Average}\\
\midrule
SoftMult & 84.25 & 81.22 & 77.57 & 72.22 & 75.79 & 74.73 & 71.28 & 61.80 & 64.69 & 50.48 & 64.46 & 40.65 & 68.26\\
SoftMult(T) & 84.24 & 81.24 & 76.88 & 71.90 & 74.90 & 74.88 & 70.37 & 59.93 & 62.86 & 52.36 & 65.78 & 43.46 & 68.23\\
NoisyOr &  82.56 & 79.69 & 76.47 & 72.81 & 75.22 & 71.84 & 71.84 & 70.81 & 77.76 & 63.74 & 68.14 & 47.47 & 71.76 \\
NoisyOr(T) & 82.56 & 79.74 & 75.87 & 72.65 & 75.69 & 73.85 & 71.13 & 69.13 & 77.19 & 64.04 & 68.52 & 49.19 & 71.63\\
\midrule
FuseNet~\cite{fusenet2016accv} & 85.41 & 80.61 & 80.98 & 73.24 & 71.59 & 69.33 & 70.85 & 49.94 & 54.55 & 3.37 & 5.58 & 4.40 & 50.89\\
SSMA ~\cite{valada2018self} & 87.35 & 82.89 & 82.58 & 73.94 & 69.88 & 65.80 & 70.90 & 33.02 & 34.15 & 51.08 & 55.35 & 42.93 & 62.49\\
\midrule
\textbf{UNO} & 82.62 & 79.69 & 77.67 & 81.24 & 80.38 & 79.57 & 78.33 & 79.72 & 79.43 & 77.55 & 79.32 & 64.80 & 78.36\\
\textbf{UNO++} &86.70	& 83.51	& 83.23 & 83.11 & 82.33 & 81.70 & 80.37	& 79.13	& 79.79	& 78.28	& 79.86	& 61.97	& \textbf{80.00}\\
\bottomrule
\end{tabular}}
\caption{\textbf{Comparison:} performance of UNO and UNO++ against other non-learning and learning baselines on Mean IoU. SoftMult(T) and NoisyOr(T) use the original temperature scaling~\cite{guo2017calibration}.}
\label{tab:comparison}
\end{table*}
\begin{table*}
\resizebox{\linewidth}{!}{%
\begin{tabular}{lrr|rrrrr|rrr}  
\toprule
&\multicolumn{2}{c}{In Distribution} & \multicolumn{5}{c}{RGB Degradation} & \multicolumn{3}{c}{Depth Degradation}\\
 \textbf{RGB}$/$\textbf{D}& 0 fog & 50 fog & 100 fog & MotionBlur & Brightness & Blackout & Impulse & Gaussian & ShotNoise & Impulse \\
\midrule
1, Ave.Temp & 1.00$/$1.00 & 1.00$/$1.00 
            & 0.97$/$1.00 & 1.00$/$1.00 & 0.82$/$1.00 & 0.03$/$1.00 & 0.05$/$1.00 
            & 1.00$/$0.88 & 1.00$/$0.93 & 1.00$/$0.75\\
2, Ave.En  & 1.00$/$1.00 & 1.00$/$1.00 
           & 0.97$/$1.00 & 0.54$/$1.00 & 0.48$/$1.00 & 0.46$/$1.00 & 0.26$/$1.00 
           & 1.00$/$0.47 & 1.00$/$0.32 & 1.00$/$1.00\\ 
\midrule
3, Ave.PreEn & 1.00$/$1.00 & 1.00$/$1.00
             &0.79$/$1.00 & 0.39$/$1.00 & 0.41$/$1.00 & 0.51$/$1.00 & 0.35$/$1.00
             &1.00$/$0.50 & 1.00$/$0.30 & 1.00$/$1.00\\
4, Ave.MI   & 1.00$/$1.00 & 1.00$/$1.00
            & 0.42$/$1.00 & 0.13$/$1.00 & 0.18$/$1.00 & 0.88$/$1.00 & 1.00$/$1.00
            & 1.00$/$0.42 & 1.00$/$0.22 & 1.00$/$1.00\\
\bottomrule
\end{tabular}}
\setlength{\belowcaptionskip}{-15pt}
\caption{\textbf{Sensitivity:} average test deviation ratio using different uncertainty metrics for in/out of distribution conditions.}
\label{tab:deviation_ratio}
\end{table*}

\section{EXPERIMENTS}
\subsection{Dataset}
We utilize the AirSim~\cite{airsim2017fsr} simulator to collect a RGB-D semantic segmentation dataset under different weather settings, such as snow and fog, that can be varied on a scale from 0 to 100. Note that a 100 setting does not correspond to complete snow whiteout or fog blackout.
All models are trained on 0 and 50 fog levels.
In order to evaluate both in- and out-of-distribution degradation, the models are tested on these original fog levels, in addition to a fog level of 100, frost, snow, and various degradation augmentation techniques investigated in~\cite{hendrycks2019robustness}. For these augmentations, we use a severity of 3 (on a scale from 0 to 5). RGB and depth frames were captured at 512x512 pixels, and following the work of~\cite{eitel2015jetmap} we use a jet mapping to pre-process our depth inputs. Overall, our dataset contains 6857 labeled training images, 472 validation images and 359 testing images.

\begin{figure}
\centering
\includegraphics[width=7.5cm]{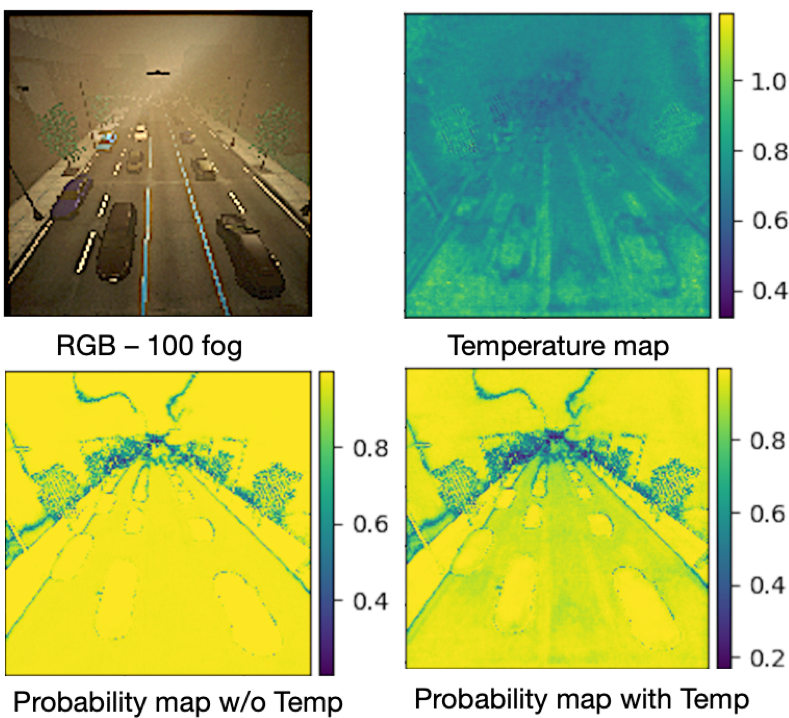}
\setlength{\belowcaptionskip}{-15pt}
\caption{\textbf{Effects of the temperature map on RGB softmax outputs}. Temperature map captures spatial uncertainty caused by fog (it is less confident in the foggy area). Smaller temperature (darker coloring) indicates lower confidence}%
\label{fig:temperature}%
\end{figure} 
\vspace{-0.1mm}
\subsection{Benchmarking}
For comparison, we consider two state-of-art learned multimodal fusion schemes: FuseNet~\cite{fusenet2016accv} and SSMA~\cite{valada2018self}.
To make the SSMA framework comparable to FuseNet and UNO, we replace the original ResNet-50 encoder \cite{he2015resnet} in SSMA with a VGG 16-layer encoder \cite{simonyan2014very}. 
To measure overall performance of our segmentation networks, we use the mean intersection-over-union (mIoU) metric~\cite{long2015fully}. 

All models are trained for a maximum of 500K iterations using Adam~\cite{kingma2014adam} with a learning rate of $10^{-5}$ and default settings, and the best on the validation set is chosen for evaluation. 
When training the temperature map and scaling parameters, we use a two-step procedure~\cite{guo2017calibration}.
First, we train our base segmentation network on our training set of semantic segmentation labels.
Then, keeping the segmentation network fixed, we train the temperature network for 100K iterations on the same training set.

\subsection{Results}\label{results}
\textbf{Fusion Performance}
In this section, we report the results of uncertainty-based Noisy-Or fusion with different deviation ratios as an ablation study in Table ~\ref{tab:ablation}, and we compare our adaptive fusion architecture to other fusion methods in~\cref{tab:comparison}.
\Cref{tab:ablation} shows that MCDO-based uncertainties (as a global indicator of the OOD ratio) do not outperform single-pass entropy. It also shows that the \textbf{min} operation can effectively choose the most sensitive uncertainty depending on the degradation.
Therefore, we use \textbf{min}(Ave.Temp,Ave.En) as the deviation ratio for our model in the following experiments because MCDO-based uncertainties require multiple passes with higher computation costs and longer runtime at inference.
\Cref{tab:comparison} compares our methods (\textbf{UNO} without and \textbf{UNO++} with SSMA as a leak expert) to other fusion methods.
The results demonstrate that our method outperforms the state-of-the-art fusion models on unseen degradations and can be easily applied to any off-the-shell multimodal fusion model to improve its performance across degradations.
Also, our baseline Noisy-Or shows better results than SSMA under unseen degradations. This improvement indicates that a multimodal expert is less robust when any of its input modalities is degraded in a manner not known \textit{a-priori}. Note also that normal temperature scaling alone does not provide additional improvement across degraded conditions.

\textbf{Temperature Maps } We qualitatively show that conventional uncertainty metrics such as predictive entropy and mutual information from multiple MCDO stochastic passes or entropy from a single deterministic pass fail at detecting uncertainty associated with degradation.
The temperature map, on the other hand, captures the spatial degradation as shown in Fig. \ref{fig:temperature}.
As shown in table~\ref{tab:deviation_ratio}, when the temperature map is used as a global scaling deviation ratio, it is sensitive to various degradation, especially when there is \textit{Impulse noise} or \textit{blackout} degradation on the RGB or depth channel.
We posit that TempNet learns data-dependent spatial uncertainty due to spatial noise through training whereas other uncertainties extract pixel-wise statistical uncertainty based on predictive distributions. 

\textbf{Uncertainty and Degradation }
In this section, we report the average test deviation ratios calculated from all aforementioned uncertainty metrics including average temperature for in-distribution and different unseen degraded conditions. 
As shown in table~\ref{tab:deviation_ratio}, we list the average deviation ratio for in-distribution conditions (0 fog and 50 fog conditions) and various OOD degradation on RGB input or depth input respectively.
The table shows that all metrics report a deviation ratio of unity on average for in-distribution data and mostly less than unity for OOD inputs.
However, they exhibit different sensitivities.
For example, average temperature is sensitive to blackout and impulse noise, and it is not as responsive for motion blur and brightness degradation on the RGB channel.
On the other hand, MCDO uncertainties and entropy are unresponsive to \textit{Impulse Noise}, justifying our combination of uncertainties.

\section{CONCLUSION AND FUTURE WORK}
We have presented an adaptive framework for multi-modal fusion that, unlike existing fusion methods, addresses unanticipated, out-of-training degradation. We benchmark different measures of uncertainty and propose a novel uncertainty-based softmax scaling as well as a deviation ratio for combining uncertainty metrics. We also propose a new data-conditioned spatial uncertainty (\textit{TempNet}) and a simple but effective noisy-or fusion scheme that can combine an arbitrary number of modalities. Results show superior performance to existing state-of-art and extensibility for incorporating them as additional experts. Next steps include experimentation with additional uncertainty metrics and analysis of their trade-offs.

\section{Acknowledgement}
\label{sec:acknowledgement}
This work was supported by ONR grant N00014-18-1-2829.

\clearpage
\printbibliography
\end{document}